\pdfoutput=1

\documentclass[11pt]{article}

\usepackage{EMNLP2022}

\usepackage{times}
\usepackage{latexsym}
\usepackage{graphicx}
\usepackage{bbm}
\usepackage{amsfonts}
\usepackage{amsmath}

\usepackage[T1]{fontenc}

\usepackage[utf8]{inputenc}

\usepackage{microtype}

\usepackage{inconsolata}

\title{C2KD: Cross-Lingual Cross-Modal Knowledge Distillation for Multilingual Text-Video Retrieval}

\author{Andrew Rouditchenko$^1$ \hspace{1mm} Yung-Sung Chuang$^1$ \hspace{1mm} Nina Shvetsova$^2$ \hspace{1mm} Samuel Thomas$^{3,4}$ \hspace{1mm} Rogerio Feris$^{3,4}$ \\
\bf Brian Kingsbury$^{3,4}$ \quad Leonid Karlinsky$^{3,4}$ \quad David Harwath$^5$ \quad Hilde Kuehne$^{2,4}$ \quad James Glass$^1$ \\
  MIT$^1$ \hspace{1mm}
  Goethe University Frankfurt$^2$ \hspace{1mm} 
  IBM Research AI$^3$ \hspace{1mm}\\
  MIT-IBM Watson AI Lab$^4$  \hspace{1mm}
  UT Austin$^5$  \\
  \texttt{roudi@mit.edu} \\
    }

\begin{document}
\maketitle
\begin{abstract}
Multilingual text-video retrieval methods have improved significantly in recent years, but the performance for other languages lags behind English.
We propose a Cross-Lingual Cross-Modal Knowledge Distillation method to improve multilingual text-video retrieval.
Inspired by the fact that English text-video retrieval outperforms other languages, we train a student model using input text in different languages to match the cross-modal predictions from teacher models using input text in English.
We propose a cross entropy based objective which forces the distribution over the student’s text-video similarity scores to be similar to those of the teacher models.
We introduce a new multilingual video dataset, Multi-YouCook2, by translating the English captions in the YouCook2 video dataset to 8 other languages.
Our method improves multilingual text-video retrieval performance on Multi-YouCook2 and several other datasets such as Multi-MSRVTT and VATEX.
We also conducted an analysis on the effectiveness of different multilingual text models as teachers.
The code, models, and dataset are available at \url{https://github.com/roudimit/c2kd}.

\end{abstract}

\section{Introduction}

Text-video retrieval, or the task of searching for videos with text queries, is becoming increasingly important as more videos are uploaded to the internet.
Currently, most methods developed for this task are trained and evaluated with English text.
The focus of this work is to improve the performance of text-video retrieval on more languages. 

Learning a multilingual multimodal embedding space~\citep{huang2021multilingual,akula2021cross} has been useful for multilingual text-video retrieval.
Text in different languages and video are processed by separate encoders and projected into the shared embedding space, where text and video that are semantically related should be close together regardless of the language.
During inference, text queries and candidate videos are projected into the embedding space, and videos are ranked according to the similarity scores between the text and video embeddings.
These methods are trained with a cross-modal contrastive objective on video datasets with parallel text translations in multiple languages, which are often derived from the original captions in English using machine translation.
They leverage recently available multilingual models  pre-trained on many languages~\cite{devlin-etal-2019-bert,conneau-etal-2020-unsupervised} to process text in different languages with only a single encoder.

While these methods have improved multilingual text-video retrieval, the performance for English is usually higher than for other languages.
Two possible reasons are: (1) multilingual text translated from English often has errors; (2) the multilingual text models are pre-trained on large-scale text data, but there is more data for English than other languages.

To address the gap in performance between English and multilingual text-video retrieval, we propose C2KD: Cross-Lingual Cross-Modal Knowledge Distillation.
Our method trains a \textit{student} model to learn better multilingual text-video similarity scores by learning from the English text-video scores of multiple trained and frozen \textit{teachers}.
The student learns to pull together video and multilingual text embeddings by optimizing their text-video scores through the contrastive loss.
We introduce a framework where several trained and frozen teachers simultaneously process the English translations of the student’s inputs and predict English text-video scores.
Further, we propose a cross entropy based objective between the student’s multilingual text-video scores and the teachers’ English text-video scores.
This teaches the student to learn multilingual text-video scores which are more aligned with the English scores, thus improving the multilingual text-video retrieval performance.

We applied our method to three existing multilingual text-video datasets: Multi-MSRVTT~\cite{huang2021multilingual}, VATEX~\cite{wang2019vatex}, and RUDDER~\cite{akula2021cross}.
Since these datasets are mainly focused on open-domain videos, we collected the Multi-YouCook2 dataset as an extension of the YouCook2~\cite{zhou2018towards} cooking video dataset to test the model in a domain which requires more fine-grained reasoning, such as understanding specific ingredients in recipes.
Our results show that C2KD can improve the multilingual text-video retrieval performance on all datasets, despite the variety in languages, domains, and dataset sizes.

In summary, our contributions are: 
(1) We propose the C2KD method which guides a student model to learn better multilingual text-video similarity scores by learning from the text-video scores of teachers using English text translations as input. 
(2) We propose a cross entropy based objective between the student and teacher text-video similarity scores to distill the cross-modal knowledge from the teachers.
(3) We collected the Multi-YouCook2 dataset with parallel text translations in 9 languages for over 10k video clips.
(4) Our method improves the multilingual text-video performance on four datasets. We conduct an analysis on the impact of different teachers to gain further insights.
The code, models, and dataset are available at \url{https://github.com/roudimit/c2kd}.
\section{Related Work}

\begin{figure*}[t!]
    \centering
    \includegraphics[width=1.0\linewidth]{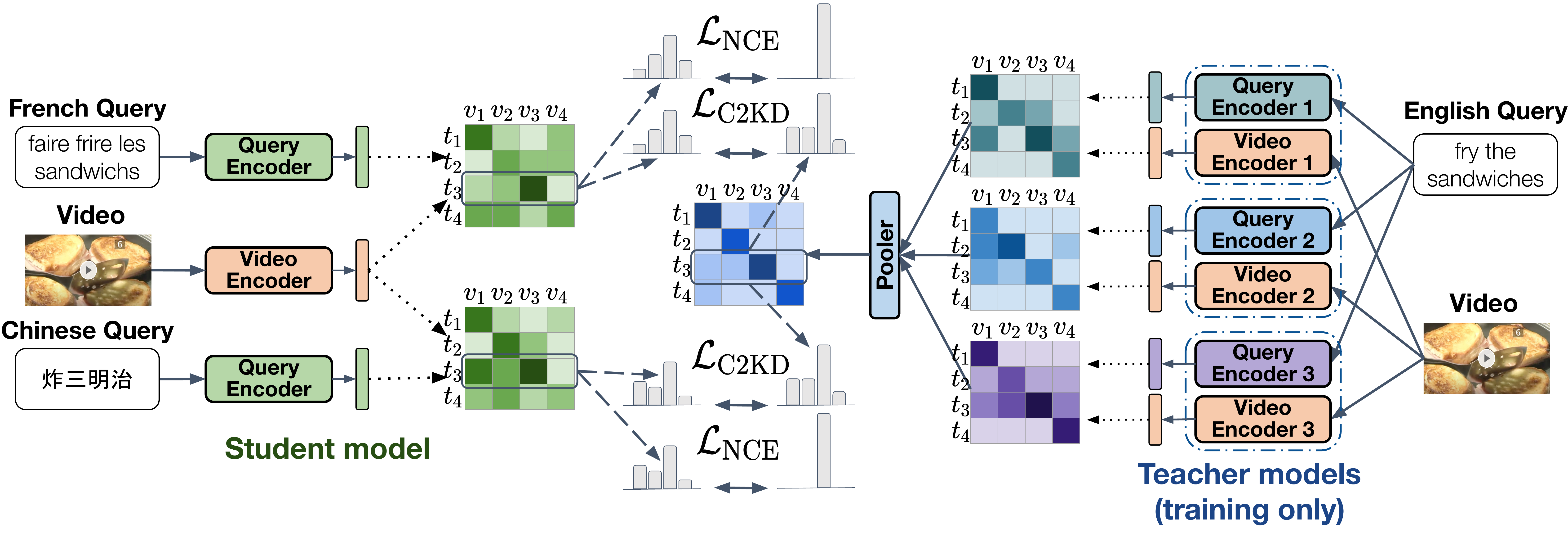}
    \caption{\textbf{Overview of C2KD.} A multilingual student model computes text-video similarity scores for a batch of video and text inputs, while teacher models process the same video and English translations.
    The student is trained with two objectives.
    $\mathcal{L}_{NCE}$ (described in Section~\ref{sec:nce_loss}) trains the model to have high text-video scores for text and video pairs using the cross entropy loss.
    $\mathcal{L}_{C2KD}$ (described in Section~\ref{sec:proposed_loss}) distills the knowledge from the teacher English text-video scores using a cross entropy loss.
    }\label{fig:diagram}
\end{figure*}

\noindent \textbf{Multilingual Text-Video Retrieval.}
Recent work introduced methods and datasets to improve multilingual text-video retrieval.
Multilingual multimodal pretraining~\cite{huang2021multilingual} demonstrated text-video retrieval in 9 languages with a single model.
They released the Multi-MSRVTT dataset by machine-translating the English text captions from the MSR-VTT video dataset~\cite{xu2016msr} into 8 other languages.
Their model is trained with a cross-modal contrastive objective to pull together the embeddings of parallel text translations and video inputs together.
In separate work, the RUDDER~\cite{akula2021cross} dataset was introduced with captions in languages spoken in India.
They propose to augment the text-video triplet loss with hard negatives which improved performance in a low-resource setting. 
We observed that performance for English text-video retrieval typically outperformed other languages, which motivated our approach.

\noindent \textbf{Multilingual Learning.}
Multilingual text-video retrieval methods rely on pre-trained multilingual text encoders to handle many languages with a single model.
MBERT~\cite{devlin-etal-2019-bert} and XLM-R~\cite{conneau-etal-2020-unsupervised} learn multilingual representations through masked language modeling.
LaBSE~\cite{feng-etal-2022-language} is instead trained to maximize the similarity of translation pairs in a shared embedding space.
In our experiments, we evaluated these different models and found LaBSE to be the best encoder for multilingual text-video retrieval.

\noindent \textbf{Cross-Lingual \& Cross-Modal Knowledge Distillation.}
Another approach for training a multilingual text model with good sentence embeddings is to distill the knowledge~\cite{hinton2015distilling} from a monolingual model.
Distill Sentence BERT~\cite{reimers-gurevych-2020-making} is initialized from XLM-R and trained to output similar multilingual embeddings to Sentence BERT~\cite{reimers-gurevych-2019-sentence} using English translations as input.
Our C2KD approach has a similar idea, but it incorporates visual context.
We use English text as input to several cross-modal teachers, and train a student to output similar text-video similarity scores using text in other languages.

Of most relevance to our work, TeachText~\cite{croitoru2021teachtext} introduced cross-modal Knowledge Distillation for English text-video retrieval.
They use teacher retrieval models with various English text embeddings and train a student to output similar text-video similarity scores with a regression loss.
Our approach has several major differences.
First, our text and models are multilingual.
Second, we enforce the teachers to use English input instead of using the same multilingual input as the students.
Third, we use a cross entropy objective between the student and teacher text-video scores instead of using a regression loss, which is more effective since it considers the context of all of the text-video pairs in the batch.
We compare our objective to theirs in Section~\ref{sec:ablation}.

Finally, some multilingual knowledge distillation methods were proposed for visual question answering based on images~\cite{raj-khan-etal-2021-towards-developing,gupta2022cvil}.

\noindent \textbf{Other Multilingual Video Datasets.}
Several multilingual video datasets are designed for other tasks, such as captioning~\cite{wang2019vatex,su-etal-2021-gem}, sentiment analysis~\cite{bagher-zadeh-etal-2020-cmu,gupta20223massiv}, moment detection~\cite{lei-etal-2021-mtvr}, audio-visual speech recognition~\cite{10.1145/3197517.3201357}, and audio-video retrieval~\cite{rouditchenko21b_interspeech}.
Instructional videos with captions from automatic speech recognition have been used for learning word embeddings~\cite{sigurdsson2020visual} and visually-guided machine translation~\cite{sanabria2018how2}. 
However, the transcriptions often have errors and can be unrelated to the visuals.
Our Multi-YouCook2 dataset contains captions which were originally written by human annotators in English~\cite{zhou2018towards}, which makes them visually relevant.

\noindent \textbf{Concurrent Work.}
~\citet{madasu2023improving} propose a similar framework to improve multilingual text-video retrieval.
However, their method uses knowledge transfer from multilingual text, while our method uses knowledge transfer from English text.
They use a separate encoder for English and multilingual text, while our final model uses a single encoder for all languages.

\section{Method}

\subsection{Text-Video Contrastive Loss}
\label{sec:nce_loss}
We handle the problem of learning multilingual text-video representations.
For simplicity, we first describe the approach for learning with English text and then explain how to extend it to more languages.
We consider a dataset $D_{en}=\{(t_i, v_i)\}_{i=1}^N$ of paired videos and English captions.
The goal of text-video retrieval is to learn text and vision models, $f(\cdot)$ and $g(\cdot)$ respectively, which output embeddings that are similar to each other when the input text caption $t_i$ and video $v_i$ are semantically related (ie. describing similar concepts), and have low similarity when they are unrelated.
In this work, we use cosine similarity by L2-normalizing the outputs of $f(\cdot)$ and $g(\cdot)$ and taking the dot-product.

The Noise-Contrastive Estimation loss (NCE)~\cite{gutmann2010noise,jozefowicz2016exploring,oord2018representation} has been commonly used to learn text-video representations~\cite{sun2019learning,rouditchenko21_interspeech}.
Given a batch of $B$ text-video pairs, let $\mathbf{S}$ be the text-video similarity matrix, with $\mathbf{S}_{ij}=f(t_i)^\top g(v_j)$.
With temperature $\tau$, the NCE loss is given as:
\begin{equation}
    \mathcal{L}_{NCE} =
    -\sum_{i=1}^B\log \frac{\exp(\mathbf{S}_{ii}/\tau)}{\sum_{k=1}^B{\exp(\mathbf{S}_{ik}/\tau)}}. \label{eq:NCE}
\end{equation}
This can be interpreted as the cross entropy loss between the distribution over normalized text-video similarity scores in $\mathbf{S}$ and the one-hot distribution.
Specifically, let $Q_{t_i}(v_j)$ be the probability that video $v_j$ matches with text $t_i$ :
\begin{equation}
    Q_{t_i}(v_j) =\frac{\exp(\mathbf{S}_{ij}/\tau)}{\sum_{k=1}^B{\exp(\mathbf{S}_{ik}/\tau)}}.
\end{equation}
The target distribution, $P_{t_i}(v_j)$, is one-hot (since the correct match for text $t_i$ is video $v_i$):
\begin{equation}
    P_{t_i}(v_j)=
    \begin{cases}
      1, & \text{if}\ i=j \\
      0, & \text{otherwise.}
    \end{cases}
\end{equation}
Given the equation for cross entropy,
\begin{equation}
 \mathcal{L}_{CE} =
 -\sum_{i=1}^B \sum_{j} P_{t_i}(v_j) \log Q_{t_i}(v_j),
 \label{eq:CE}
\end{equation}
we can see immediately that Eq.~\ref{eq:NCE} is equivalent to Eq.~\ref{eq:CE}.
In Section~\ref{sec:proposed_loss}, we introduce an additional cross entropy based objective between a new target distribution $P'_{t_i}(v_j)$ and $Q_{t_i}(v_j)$.

To extend this to a dataset of videos paired with captions in $L$ languages, ie. $D_{multi}=\{(t_i^1, t_i^2, \dots, t_i^L, v_i)\}_{i=1}^N$, we compute a text-video similarity matrix for each language, ie. $\mathbf{S}^l$, where $\mathbf{S}^l_{ij}=f(t_i^l)^\top g(v_j)$.
Then we apply $\mathcal{L}_{NCE}$ to each matrix and take the sum of the losses.
This pulls together the embeddings of videos and their paired captions in different languages.

During inference, $f$ and $g$ are used to encode text and video inputs.
For a given text query, videos are ranked by their cosine similarity to the text. 

\subsection{C2KD Method}
Although $\mathcal{L}_{NCE}$ can be used to learn multilingual text-video representations, the performance for English text-video retrieval is usually higher than for other languages, as mentioned in the introduction.
This implies that of all the languages, the English text-video similarity scores are most accurate.
Our key idea is to use the English text-video similarity scores to improve the scores for other languages.

The method is illustrated in Figure~\ref{fig:diagram}.
We first train $M$ teacher models using $D_{multi}$ and $\mathcal{L}_{NCE}$, and then freeze their parameters.
The teacher models have the same architecture, except the text encoders are different so that complementary information from different models can be used.
Next, we begin training a student model with $D_{multi}$ and $\mathcal{L}_{NCE}$. 
For each batch of video and multilingual text, the teachers are simultaneously provided with the video and English translations as input.
Each teacher produces an English text-video similarity matrix.
We apply a pooler function $\Psi: \mathbb{R}^{M \times B\times B} \rightarrow \mathbb{R}^{B\times B}$ to the $M$ teacher similarity matrices to get a single similarity matrix $\mathbf{S}'$, where $\mathbf{S'}_{ij}$ is the similarity score at row $i$ and column $j$.
In our experiments, we experimented with different pooler functions such as mean, max, and min.
We train the student with $\mathcal{L}_{NCE}$ and a 2nd objective, $\mathcal{L}_{C2KD}$ (introduced in Section~\ref{sec:proposed_loss}), which encourages the student's text-video similarity scores from captions in different languages to be similar to the teacher English text-video scores in $\mathbf{S}'$.
Note that only the student model is used during inference.

\subsection{Knowledge Distillation Objective}
\label{sec:proposed_loss}
We introduce a distillation objective that encourages the student's multilingual text-video similarity scores to be similar to the teacher English text-video scores in $\mathbf{S}'$.
The main idea is that instead of using the one-hot distribution $P_{t_i}(v_j)$ in $\mathcal{L}_{NCE}$, we use a new distribution $P'_{t_i}(v_j)$ obtained from the teacher English text-video scores in $\mathbf{S}'$.
Specifically, let $P'_{t_i}(v_j)$ be the probability that video $v_j$ matches with text $t_i$ :
\begin{equation}
    P'_{t_i}(v_j) = \frac{\exp(\mathbf{S'}_{ij}/\tau)}{\sum_{k=1}^B{\exp(\mathbf{S'}_{ik}/\tau)}}
\end{equation}
We apply the cross entropy loss between $P'_{t_i}(v_j)$ (generated by the teacher English text-video similarity scores) and $Q_{t_i}(v_j)$ (generated by the student multilingual text-video similarity scores): 
\begin{equation}
 \mathcal{L}_{C2KD} =
 -\sum_{i=1}^B \sum_{j}P'_{t_i}(v_j) \log Q_{t_i}(v_j),
\end{equation}
Note that the temperature $\tau$ in $\mathcal{L}_{C2KD}$ is controlled separately to the one in $\mathcal{L}_{NCE}$.
We apply $\mathcal{L}_{C2KD}$ to each of the student text-video similarity matrices using text in different languages and take the sum of the losses.
The final objective is given by:
\begin{equation}
 \mathcal{L} = \alpha \mathcal{L}_{NCE} + (1 - \alpha) \mathcal{L}_{C2KD}
\end{equation}
where $\alpha$ is a balance hyperparameter.

The difference between $\mathcal{L}_{NCE}$ and $\mathcal{L}_{C2KD}$ is the target distribution; the former uses a one-hot distribution while the latter uses soft-labels produced by the teachers.
$\mathcal{L}_{NCE}$ makes rigid assumptions about which captions are similar to which video clips (only paired examples should match), whereas $\mathcal{L}_{C2KD}$ enables the model to have leeway in assigning higher scores to pairs which are not ground-truth pairs, but still have some semantic similarity.
Also, $\mathcal{L}_{C2KD}$ shares the same cross entropy objective as the original KD~\cite{hinton2015distilling}, but it is more technically advanced since it distills teacher cross-modal matrices instead of just the logits from uni-modal encoders.
Further, our cross-modal distillation is consistent with the retrieval task.

Others have also observed that the contrastive loss may be too strict in a cross-modal setting and have proposed complementary objectives such as captioning~\cite{patrick2020support} and clustering~\cite{chen2021multimodal,liu-etal-2022-cross}.
However, to the best of our knowledge, it is novel to use the contrastive loss in a cross-modal setting with target distributions generated from the text-video similarity scores of several teachers.

\section{Experiments}

\subsection{Datasets}

\noindent \textbf{Multi-MSRVTT}~\cite{huang2021multilingual} is a multilingual version of the MSRVTT~\cite{xu2016msr} video dataset.
The video categories are general, such as ``sports'' and ``vehicles.''
The original dataset contains 10k videos from YouTube, each annotated with 20 captions in English.
The captions were translated to 8 other languages with machine translation.
We followed the setup in prior work~\cite{huang2021multilingual} and used a training set of 6.5k videos, validation set of 497 videos, and test set of 1k videos.

\begin{figure}[t!]
    \centering
    \includegraphics[width=1.0\linewidth]{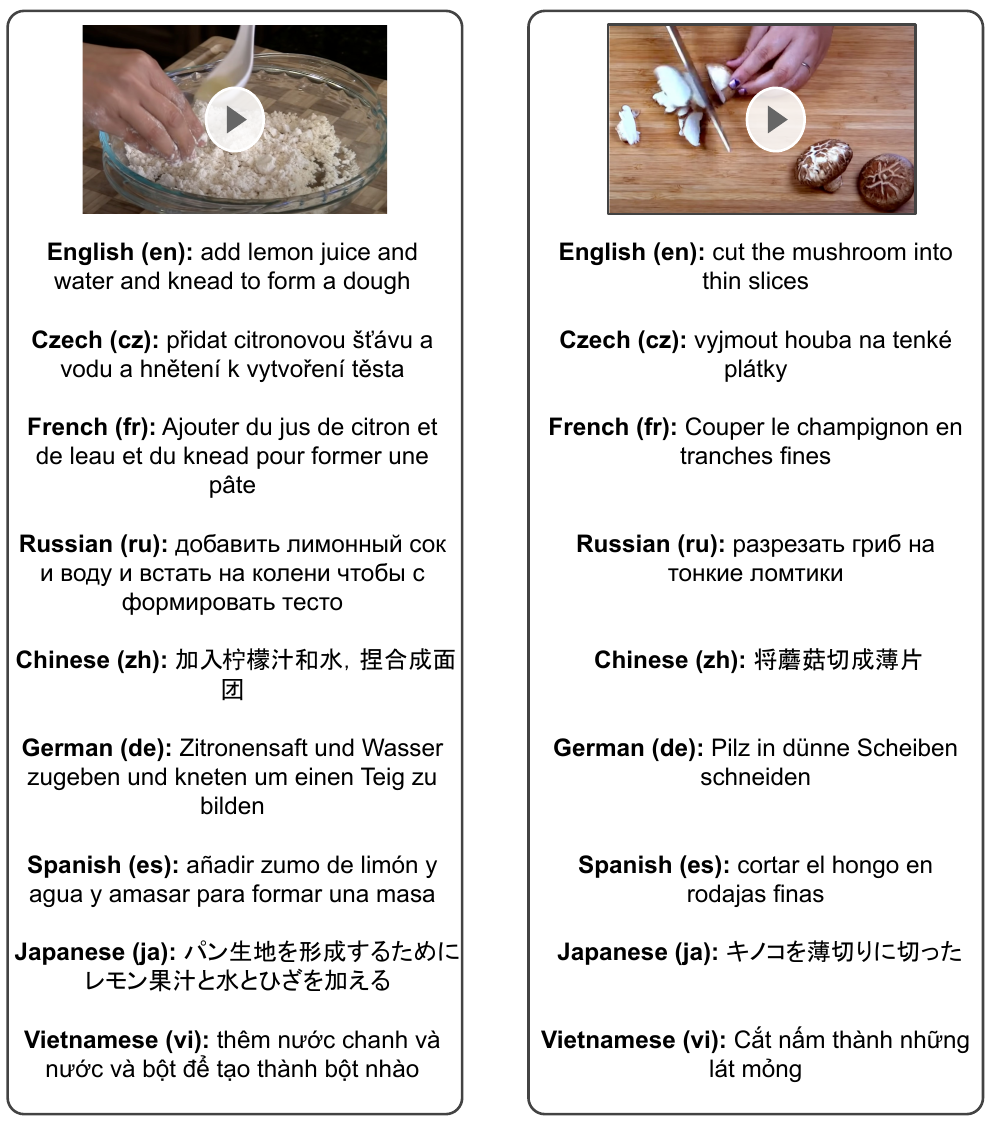}
    \caption{\textbf{Multi-YouCook2 sample video clips and multilingual captions.} 
    }\label{fig:dataset}
\end{figure}

\noindent \textbf{Multi-YouCook2} is our multilingual extension of the YouCook2~\cite{zhou2018towards} video dataset.
The original dataset contains 2k cooking videos from YouTube.
The video categories are about recipes, such as ``spaghetti and meatballs.''
Each video was segmented into smaller clips containing recipe steps and annotated with text captions of the recipe steps in English.
Inspired by the procedure to collect Multi-MSRVTT~\cite{huang2021multilingual}, we translated the captions to 8 other languages using machine translation.
Sample clips and captions are shown in Figure~\ref{fig:dataset}.
Following the setup in prior English text-video work~\cite{miech2019howto100m}, we used 9,586 training clips and 3,350 evaluation clips.

\noindent \textbf{VATEX}~\cite{wang2019vatex} contains videos each with 10 English and 10 Chinese captions.
The videos were selected from an action classification dataset~\cite{kay2017kinetics}.
Following prior work~\cite{huang2021multilingual}, we use the official training set of 26k videos and split the validation set equally into 1.5k validation and 1.5k test videos.
Note that we made our own split since theirs was not released, and we will release our split.

\noindent \textbf{RUDDER}~\cite{akula2021cross} contains instructional videos with captions in languages spoken in India.
This dataset is by far the smallest and could be considered a low-resource multilingual video dataset.
The dataset contains English captions but the original work did not use them.
Therefore, we created a split of 2.2k training clips and 1k evaluation clips, where each clip in the evaluation set has a caption in English and 3 other languages.

\subsection{Implementation Details}
For the student text model $f$, we use LaBSE~\cite{feng-etal-2022-language}.
We discuss the teacher text models in Section~\ref{sec:ablation}.
For the video model $g$, we first extract features from CLIP ViT-B/32~\cite{radford2021learning} at 1 FPS and process them with a 2-layer Transformer~\cite{vaswani2017attention}.
Due to GPU memory limitations, we do not update the weights of the CLIP model.
We set $\tau$ in $\mathcal{L}_{NCE}$ to 0.05 and $\tau$ in $\mathcal{L}_{C2KD}$ to 0.1.
We found the best pooler function $\Psi$ and balance $\alpha$ to be different for each dataset.
We specify the values and discuss other hyperparameters in the Appendix.
\subsection{Experimental Setup}
We use the standard R@K metrics (recall at rank K, higher is better).
All of our reported results are the average of three runs.
Note that random chance performance is different on each dataset due to varying evaluation set size.
In the zero-shot setting, models are trained on English text-video pairs only and evaluated using captions in all languages.
In the translate-train setting, the models are trained on text-video pairs in all languages.
Note that C2KD is only applicable to the translate-train setting since it requires multilingual text during training.

\begin{figure}[t!]
    \centering
    \includegraphics[width=1.0\linewidth]{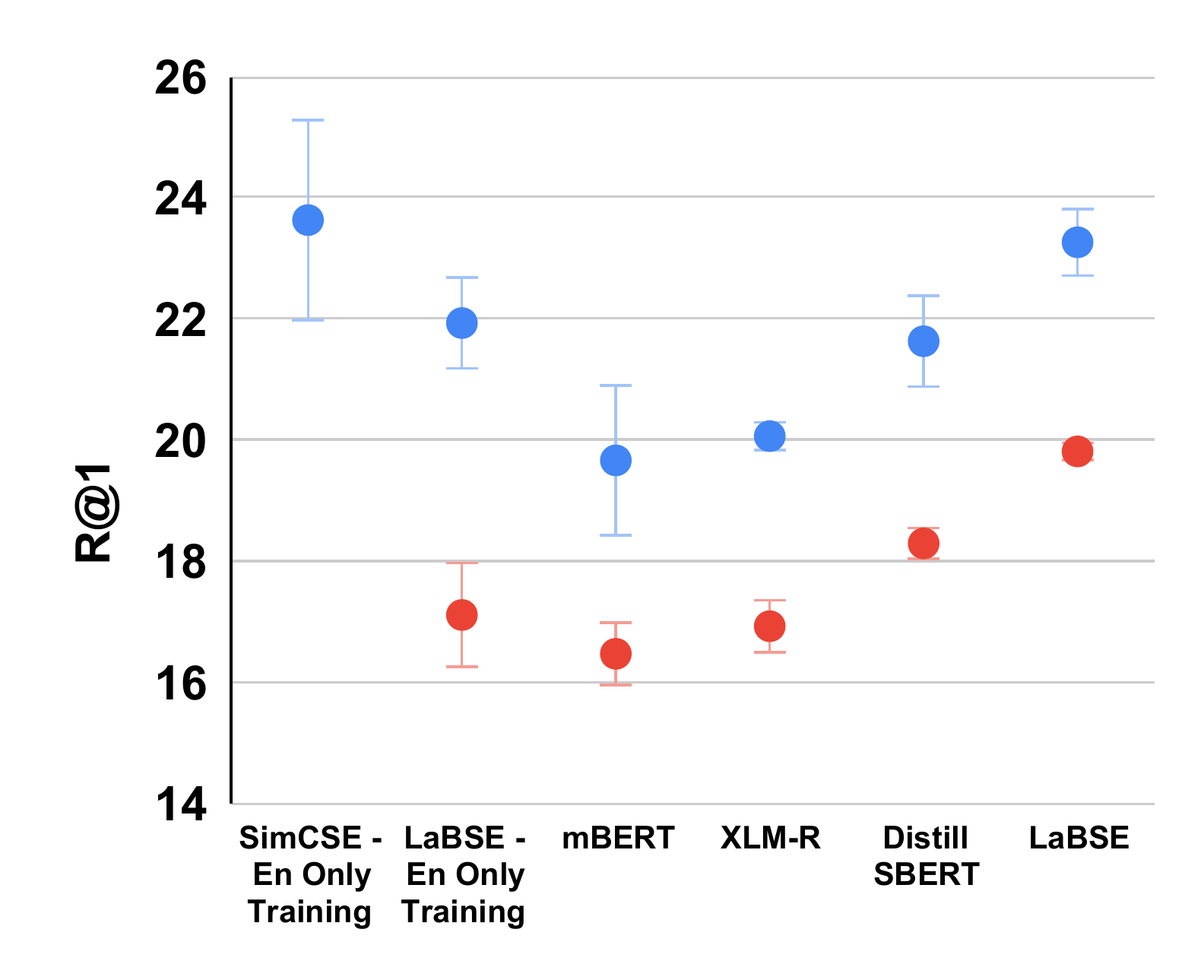}
    \caption{\textbf{Text encoder study.} 
    The blue dots show text-video retrieval performance using English.
    The red dots show the average text-video retrieval performance using 9 languages.
    }\label{fig:students_msrvtt}
\end{figure}

\begin{figure}[t!]
    \centering
    \includegraphics[width=1.0\linewidth]{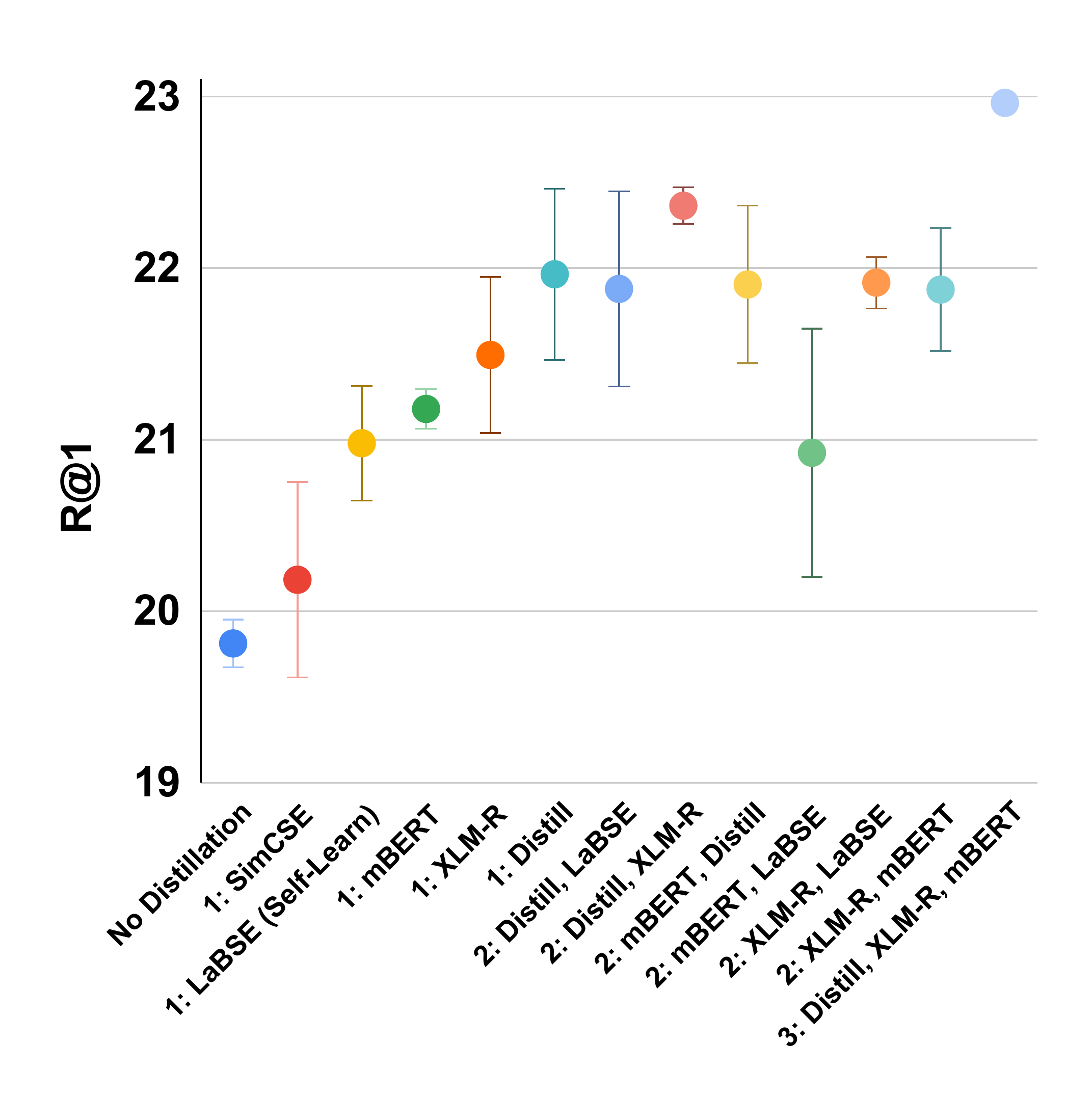}
    \caption{\textbf{Ablation on the number of teachers.}
    Using LaBSE as the student, we applied different combinations of models as teachers.
    }\label{fig:teachers_msrvtt}
\end{figure}

\begin{figure}[t!]
    \centering
    \includegraphics[width=0.75\linewidth]{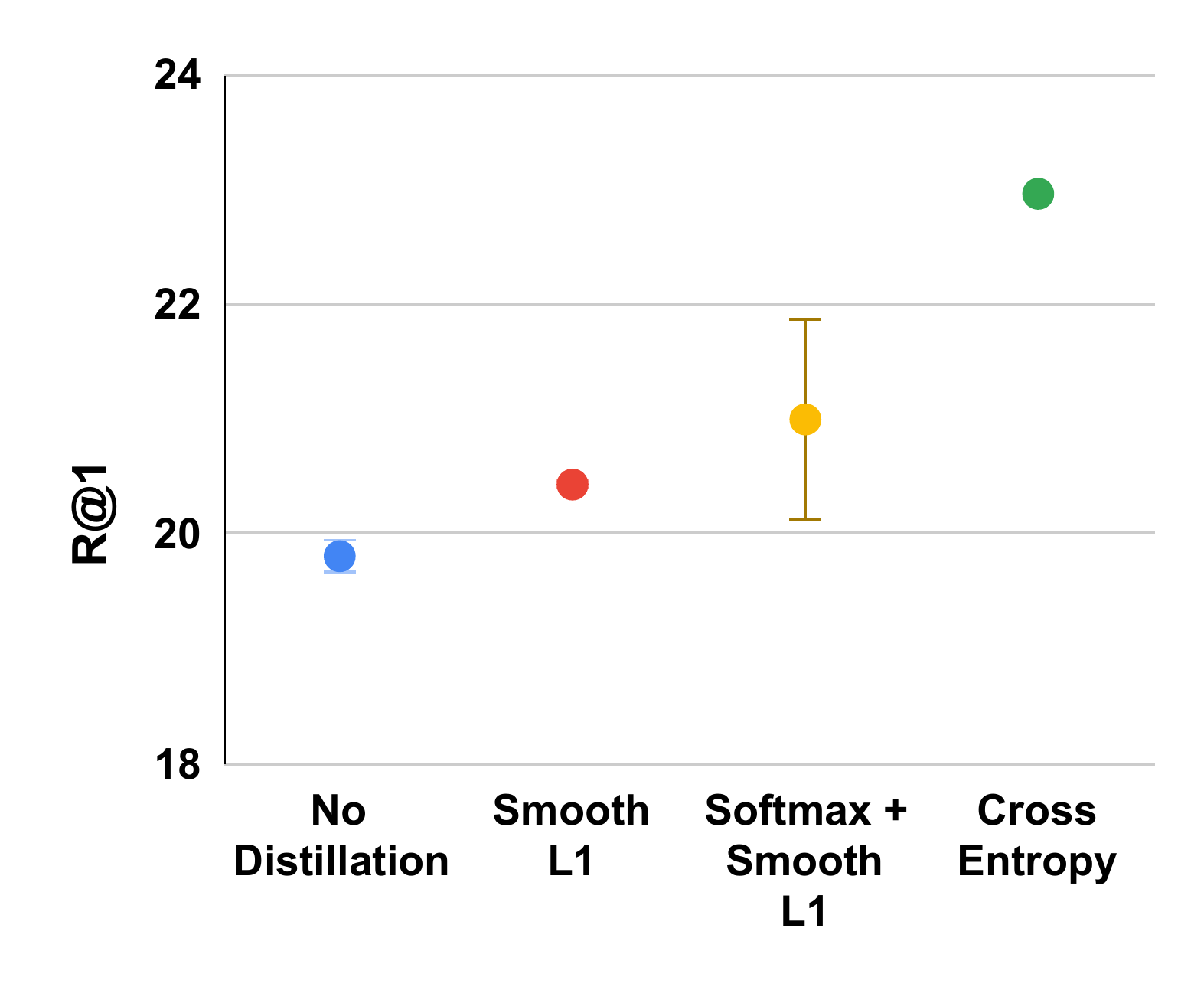}
    \caption{\textbf{Ablation on the Knowledge Distillation objective.}}\label{fig:kd_objective}
\end{figure}

\begin{figure}[t!]
    \centering
    \includegraphics[width=0.75\linewidth]{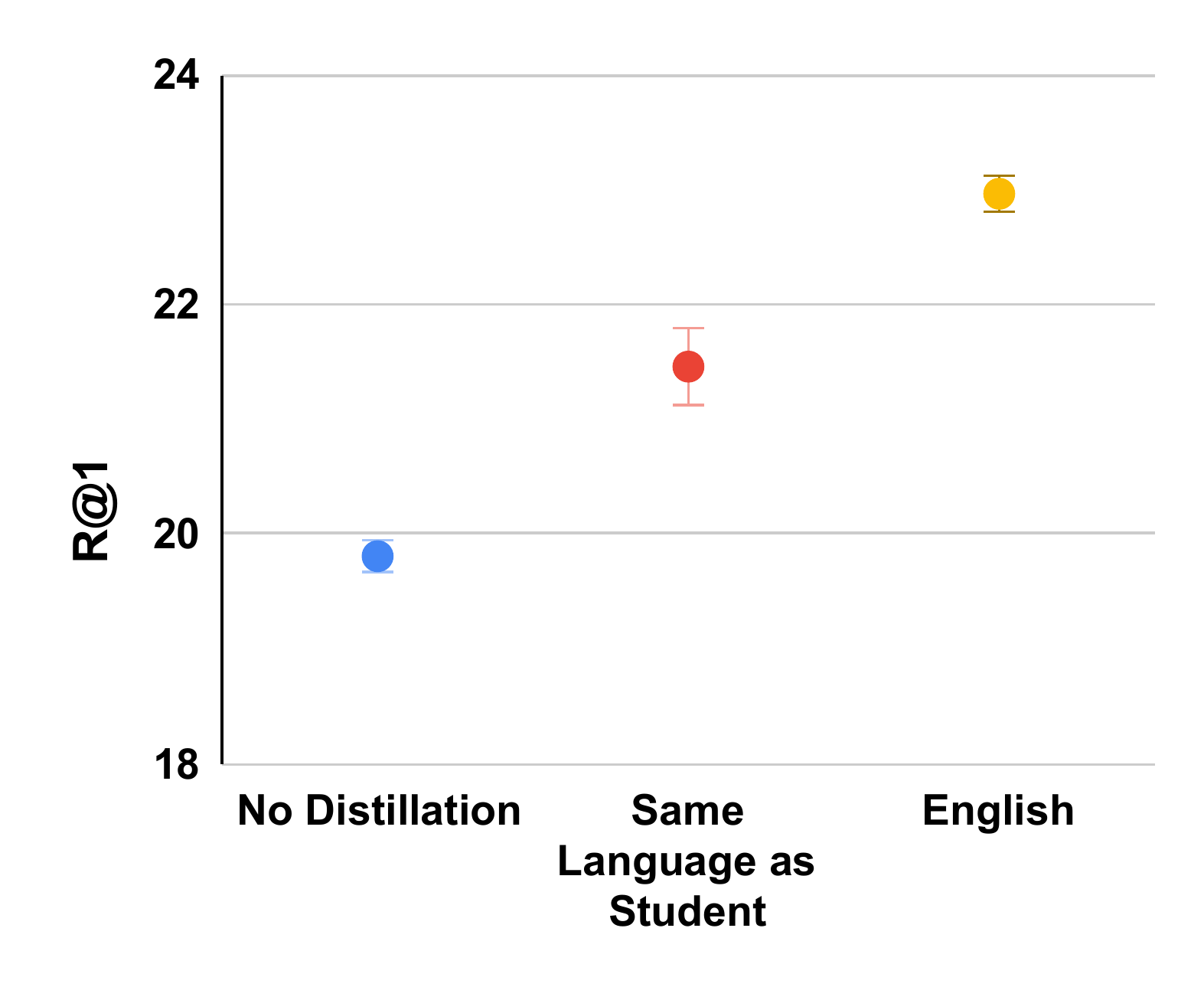}
    \caption{\textbf{Ablation on the language used with the teachers.}}\label{fig:kd_application}
\end{figure}

\begin{table*}[th!]
\small
\centering
 \begin{tabular}{llcccccccccc }
      \hline
        Method & Set. & $en$ & $de$ & $fr$ & $cs$ & $zh$ & $ru$  & $vi$ & $sw$ & $es$ & Avg$\uparrow$ \\
        \hline \hline 
        Random Chance & N/A & 0.1 & 0.1 & 0.1 & 0.1 & 0.1 & 0.1 & 0.1 & 0.1 & 0.1 & 0.1 \\
        \hline 
        PO\textsuperscript{\textdagger} \cite{akula2021cross} & ZS & 17.7 & 15.1 & 14.8 & 13.0 & 11.6 & 12.6 & 7.1 & 4.9 & 15.6 & 12.5 \\
        MMP \cite{huang2021multilingual} & ZS & \textbf{23.8} & \textbf{19.4} & \textbf{20.7} & \textbf{19.3} & \textbf{18.2} & \textbf{19.1} &  8.2 & 8.4 & 20.4 & \textbf{17.5} \\
        NCE & ZS & 21.9 & 18.9 & 18.7 & 18.2 & 16.3 & 17.5 & \textbf{9.1} & \textbf{12.8} & \textbf{20.5} & 17.1 \\
        \hline
        PO\textsuperscript{\textdagger} \cite{akula2021cross} & TT & 17.0 & 17.0 & 17.2 & 16.1 & 14.6 & 16.0 & 8.6 & 11.5 & 16.8 & 15.0 \\
        MMP \cite{huang2021multilingual} & TT & 23.1 & 21.1 & 21.8 & 20.7 & 20.0 & \textbf{20.5} & 10.9 & 14.4 & \textbf{21.9} & 19.4\\
        NCE & TT & \textbf{23.3} & \textbf{21.1} & \textbf{22.3} & \textbf{20.9} & \textbf{20.3} & 19.6 & \textbf{12.1} & \textbf{17.2} & 21.5 & \textbf{19.8}\\
        \hline
        C2KD (ours) & TT & \textbf{26.4} & \textbf{24.7} & \textbf{25.4} & \textbf{24.0} & \textbf{23.4} & \textbf{23.1} & \textbf{13.6} & \textbf{20.3} & \textbf{25.5} & \textbf{23.0}\\
        \hline
    \end{tabular}
    \caption{\textbf{Multilingual text-video retrieval on Multi-MSRVTT (R@1).} \textdagger: our implementation, Set.=Setting, ZS=Zero-Shot (trained on English text-video only), TT=Translate-Train (trained on text-video in all languages).
    }\label{tab:msrvtt}
\end{table*}

\begin{table*}[th!]
\small
\centering
 \begin{tabular}{llcccccccccc }
      \hline
        Method & Set. & $en$ & $de$ & $fr$ & $cs$ & $zh$ & $ru$  & $vi$ & $ja$ & $es$ & Avg$\uparrow$ \\
        \hline \hline 
        Random Chance & N/A & 0.03 & 0.03 & 0.03 & 0.03 & 0.03 & 0.03 & 0.03 & 0.03 & 0.03 & 0.03 \\
        \hline
        PO\textsuperscript{\textdagger}~\cite{akula2021cross} & ZS & 10.1 & 2.5 & 2.7 & 2.1 & 1.4 & 1.6 & 2.2 & 1.2 & 2.3 & 2.9 \\
        MMP\textsuperscript{\textdagger}~\cite{huang2021multilingual} & ZS & 12.7 & 3.7 & 3.3 & 2.7 & 2.0 & 2.5 & 2.3 & 1.8 & 2.4 & 3.7 \\
        NCE & ZS & \textbf{14.4} & \textbf{7.0} & \textbf{6.4} & \textbf{5.1} & \textbf{3.5} & \textbf{4.7} & \textbf{5.0} & \textbf{2.7} & \textbf{6.3} & \textbf{6.1} \\
        \hline 
        PO\textsuperscript{\textdagger}~\cite{akula2021cross} & TT & 10.0 & 9.1 & 9.1 & 8.6 & 6.7 & 9.0 & 6.3 & 7.5 & 9.1 & 8.4 \\
        MMP\textsuperscript{\textdagger}~\cite{huang2021multilingual} & TT & 11.3 & 10.4 & 10.6 & 10.1 & 8.3 & 9.3 & 8.4 & 9.1 & 10.4 & 9.8 \\
        NCE & TT & \textbf{14.9} & \textbf{13.1} & \textbf{13.0} & \textbf{12.1} & \textbf{9.6} & \textbf{12.1} & \textbf{10.9} & \textbf{10.0} & \textbf{13.2} & \textbf{12.1} \\
        \hline
        C2KD (ours) & TT & \textbf{15.5} & \textbf{14.0} & \textbf{13.9} & \textbf{12.8} & \textbf{10.4} & \textbf{13.1} & \textbf{11.4} & \textbf{11.3} & \textbf{14.1} & \textbf{12.9} \\
        \hline
    \end{tabular}
    \caption{\textbf{Multilingual text-video retrieval on Multi-YouCook2 (R@1).} \textdagger: our implementation, Set.=Setting, ZS=Zero-Shot (trained on English text-video only), TT=Translate-Train (trained on text-video in all languages).
    }\label{tab:youcook2}
\end{table*}

\begin{table}[th!]
\small
\centering
\resizebox{0.48\textwidth}{!}{%
\setlength{\tabcolsep}{2.0pt}
\begin{tabular}{llccccccccc}
\hline
    & & \multicolumn{3}{c}{English ($en$)} & \multicolumn{3}{c}{Chinese ($zh$)} \\
    Model  & Set. & R@1 & R@5 & R10 & R@1 & R@5 & R@10  \\ \hline\hline
    Random Chance & N/A & 0.07 & 0.33 & 0.67 & 0.07 & 0.33 & 0.67 \\
    \hline
    MMP \cite{huang2021multilingual} & ZS & \textbf{44.4} & 80.5 & 88.7 & 29.7 & 63.2 & 75.5 \\
    MMP \cite{huang2021multilingual} & TT &  44.3 & \textbf{80.7} & \textbf{88.9} & \textbf{40.5} & \textbf{76.4} & \textbf{85.9} \\
    \hline \hline
    PO\textsuperscript{\textdagger} \cite{akula2021cross} & ZS & 37.7 & 77.0 & 87.7 & 25.7 & 57.3 & 72.5 \\
    MMP\textsuperscript{\textdagger} \cite{huang2021multilingual} & ZS & 39.9 & 79.1 & 89.3 & 26.9 & 60.4 & 75.3 \\
    NCE & ZS & \textbf{42.0} & \textbf{81.0} & \textbf{90.6} & \textbf{28.0} & \textbf{63.4} & \textbf{75.6} \\
    \hline
    PO\textsuperscript{\textdagger} \cite{akula2021cross} & TT & 37.5 & 77.1 & 88.2 & 33.2 & 70.9 & 83.9 \\
    MMP\textsuperscript{\textdagger} \cite{huang2021multilingual} & TT & 41.3 & 78.9 & 88.8 & 34.1 & 74.3 & 85.2 \\
    NCE & TT & \textbf{42.6} & \textbf{81.0} & \textbf{90.6} & \textbf{38.0} & \textbf{75.4} & \textbf{88.0} \\
    \hline
    C2KD (ours) & TT &  \textbf{43.1} & \textbf{82.1} & \textbf{91.5} & \textbf{39.6} & \textbf{77.0} & \textbf{88.6} \\
    \hline
\end{tabular}}
\caption{\textbf{Multilingual text-video retrieval on VATEX.} Upper and lower halves separated due to different test splits. \textdagger: our implementation, Set.=Setting, ZS=Zero-Shot (trained on English text-video only), TT=Translate-Train (trained on text-video in all languages).} \label{tab:vatex}
\vspace{-1.0em}
\end{table}

\begin{table*}[th!]
\small
\centering
\setlength\tabcolsep{3.0pt}
\begin{tabular}{lrcccccccccccc}
\hline
 & Set. & \multicolumn{3}{c}{English ($en$)} & \multicolumn{3}{c}{Hindi ($hi$)} & \multicolumn{3}{c}{Kannada ($kn$)} & \multicolumn{3}{c}{Marathi ($mr$)}\\
Model & & R@1 & R@5 & R10 & R@1 & R@5 & R@10 & R@1 & R@5 & R@10 & R@1 & R@5 & R@10 \\ \hline\hline
Random Chance & N/A & 0.1 & 0.5 & 1.0 & 0.1 & 0.5 & 1.0 & 0.1 & 0.5 & 1.0 & 0.1 & 0.5 & 1.0 \\
\hline
PO\textsuperscript{\textdagger} \cite{akula2021cross} & ZS & 4.2 & 12.7 & 17.7 & 1.8 & 6.6 & 10.2 & 1.8 & 5.7 & 9.1 & 2.8 & 7.7 & 12.8 \\
MMP\textsuperscript{\textdagger} \cite{huang2021multilingual} & ZS & 4.8 & 13.5 & 20.9 & 1.9 & 6.4 & 10.8 & 1.5 & 4.4 & 7.6 & 2.3 & 7.4 & 11.5 \\
NCE & ZS & \textbf{5.1} & \textbf{14.4} & \textbf{21.6} & \textbf{2.3} & \textbf{8.0} & \textbf{12.5} & \textbf{2.1} & \textbf{6.6} & \textbf{11.5} & \textbf{2.8} & \textbf{8.5} & \textbf{13.5} \\
\hline
PO\textsuperscript{\textdagger} \cite{akula2021cross} & TT & 4.3 & 13.4 & 18.7 & 2.9 & 10.2 & 15.4 & 2.3 & 7.9 & 13.2 & 3.7 & 11.5 & 17.0 \\
MMP\textsuperscript{\textdagger} \cite{huang2021multilingual} & TT & 5.1 & 13.4 & 19.3 & 3.1 & 10.5 & 15.2 & 3.0 & 7.8 & 12.1 & 2.9 & 11.5 & 17.6 \\
NCE & TT & \textbf{5.6} & \textbf{16.6} & \textbf{23.4} & \textbf{4.2} & \textbf{11.8} & \textbf{16.8} & \textbf{3.7} & \textbf{10.7} & \textbf{15.8} & \textbf{5.1} & \textbf{14.1} & \textbf{19.9} \\
\hline
C2KD (ours) & TT & \textbf{6.3} & \textbf{16.8} & \textbf{25.9} & \textbf{4.3} & \textbf{13.2} & \textbf{19.4} & \textbf{4.4} & \textbf{12.4} & \textbf{17.8} & \textbf{5.1} & \textbf{14.8} & \textbf{22.3} \\
\hline
\end{tabular}
\caption{\textbf{Multilingual text-video retrieval on RUDDER.} \textdagger: our implementation, Set.=Setting, ZS=Zero-Shot (trained on English text-video only), TT=Translate-Train (trained on text-video in all languages).} \label{tab:rudder}
\vspace{-1.0em}
\end{table*}

\subsection{Ablation Studies and Analysis}
\label{sec:ablation}
In this section, we conduct an analysis in order to justify our choices for the student and teacher models, as well as the design of our distillation setup.
All of the studies were done on Multi-MSRVTT.
The bars in the figures report the standard deviation of three runs.

\noindent \textbf{Text encoders.}
We compare different text encoders in Figure~\ref{fig:students_msrvtt} when trained for text-video retrieval ($\mathcal{L}_{NCE}$ only, $\alpha$=1).
LaBSE and Distill SBERT outperformed mBERT and XLM-R, which are not trained with sentence level objectives.
When trained with multilingual captions, LaBSE's performance on English is comparable to SimCSE's, a recent English-only sentence embedding model~\cite{gao-etal-2021-simcse}.
Finally, LaBSE’s performance across all languages, including English, improved when trained on multilingual captions. 
Given that LaBSE is the strongest multilingual model, we use it as our student text encoder.

\begin{figure}[t!]
    \centering
    \includegraphics[width=1.0\linewidth]{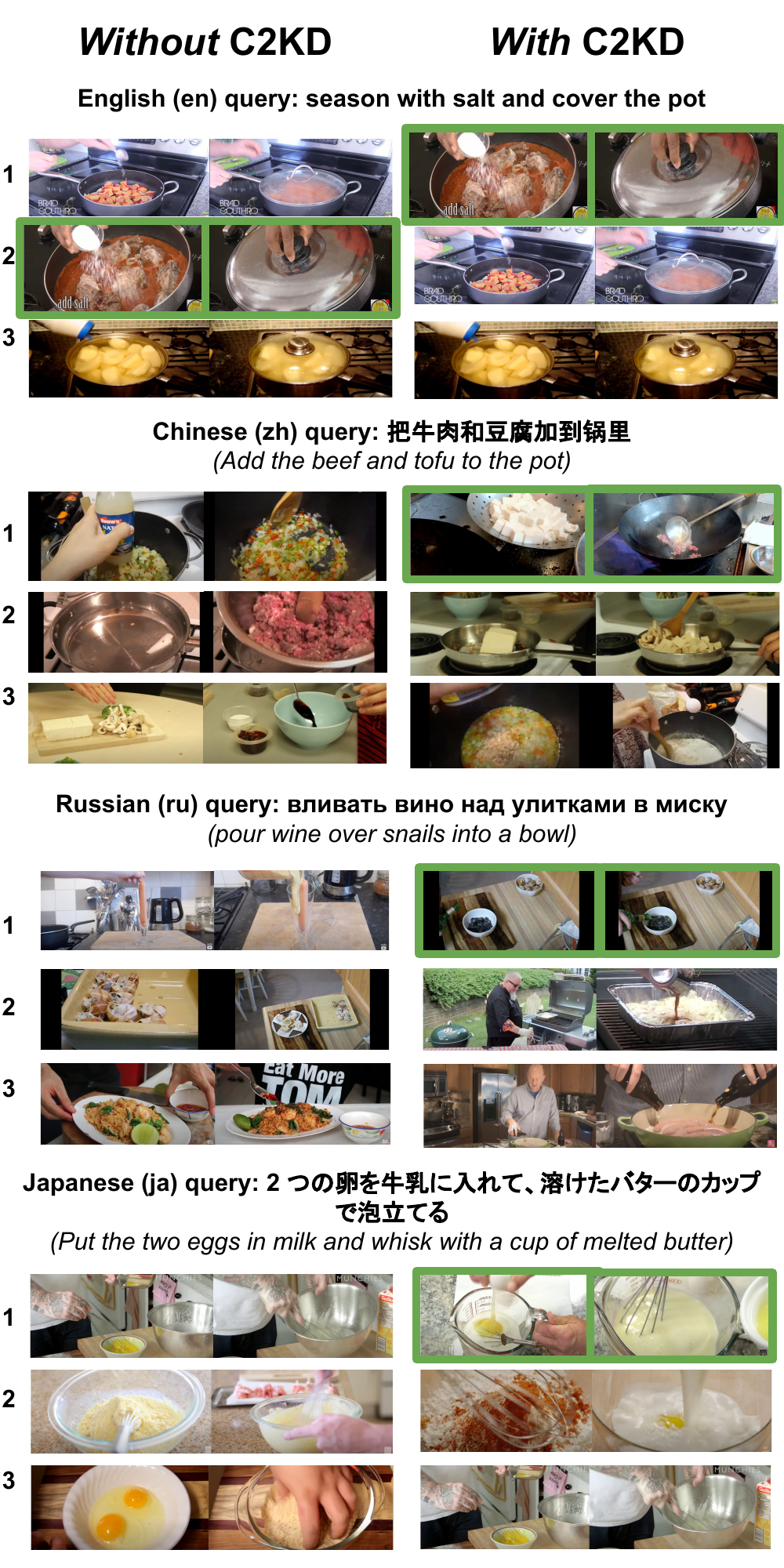}
    \caption{\textbf{Qualitative text-video retrieval results on Multi-YouCook2.} Left: results for NCE baseline method without C2KD. Right: results with C2KD. Videos shown as 2 frames. Top 3 results for each query are shown with the correct match highlighted in green.
    }\label{fig:qual}
\end{figure}

\noindent \textbf{Teacher models.}
In Figure~\ref{fig:teachers_msrvtt}, we show the performance of our C2KD method with one, two, and three teachers.
With one teacher, we found that SimCSE was the worst teacher, which was surprising considering its strong English-only performance.
However, recent work has shown that the best teacher might not be the strongest model~\cite{gong2022cmkd}.
Using LaBSE as its own teacher is feasible, but it is better to use a different model as the teacher.
Distill SBERT was the best teacher, which is reasonable considering it is the most similar to LaBSE in using a sentence-level objective.
With two teachers, we found that any combination of Distill SBERT, mBERT, and XLM-R could improve the performance over any individual teacher.
However, including LaBSE either didn't help the performance or made it worse.
Given these results, we used Distill SBERT, mBERT, and XLM-R as the final set of teachers, which obtained the best results.

\noindent \textbf{Knowledge Distillation objective.}
We compare distillation objectives between the student and teacher text-video similarity scores in Figure~\ref{fig:kd_objective}.
For English text-video retrieval, TeachText~\cite{croitoru2021teachtext} proposed to regress the teacher text-video scores using a Smooth L1 Loss.
We found that this objective could only give a minor improvement over the baseline without distillation.
While the TeachText approach considers each text-video score independently, our proposed $\mathcal{L}_{C2KD}$ loss instead considers the context of all the text-video scores by normalizing them with softmax and applying the cross entropy loss.
As shown in Figure~\ref{fig:kd_objective}, this significantly outperforms the regression based method.
To gain further insight, we tried an intermediate approach of combining softmax normalization and Smooth L1 Loss, which performed only slightly better than Smooth L1 loss.
This shows that it is essential to use a loss such as cross entropy which considers the distribution over the text-video scores instead of treating them independently.

\noindent \textbf{Teacher language.}
In Figure~\ref{fig:kd_application}, we compare the results when different languages are used by the teachers.
Using the same multilingual text as input to the student and teachers improves the results over no distillation, likely due to the complementary information provided by different text encoders.
However, our proposed method of using English with the teachers performs better.
This result matches our intuition that English should be the best language to use with the teachers since English text-video retrieval is typically higher than other languages.

\subsection{Main Results}
With the best student (LaBSE) and teacher models (Distill SBERT, mBERT, and XLM-R) at hand, we tested C2KD on four datasets. 
For comparison, we also implemented the baselines~\cite{huang2021multilingual,akula2021cross} since their code was not released.
The ``NCE'' method corresponds to our baseline without distillation ($\mathcal{L}_{NCE}$ only, $\alpha$=1).
We applied C2KD to this method.

Table~\ref{tab:msrvtt} shows the multilingual text-video retrieval results on Multi-MSRVTT.
C2KD improves performance across languages, with average R@1 improving from 19.8 to 23.0 (+16.2\% relative).
The largest improvement is on Spanish ($es$), from 21.5 to 25.5 (+18.6\% relative). 
Our method also improves the performance on English.
Finally, we note that while performance for Vietnamese ($vi$) improved, it is still much lower than for other languages. 
We manually expected the captions and found the translations to be poor with unrelated symbols such as musical notes inserted.

Table~\ref{tab:youcook2} shows the results on Multi-YouCook2.
The performance across languages is similar, which suggests that the quality translation is consistent.
Applying our C2KD method to the baseline, we see improvements for all languages.
The average R@1 improves from 12.1 to 12.9 (+6.6\% relative).
The largest improvement is on Japanese, from 10.1 to 11.3 (+11.9\% relative).

Table~\ref{tab:vatex} shows the results on VATEX.
The retrieval performance is generally higher than on the other datasets, which could be attributed to the large training set.
Nonetheless, C2KD can improve the performance for both English and Chinese in all metrics.
Chinese R@1 is improved from 38 to 39.6 (+4.2\% relative).

Table~\ref{tab:rudder} shows the results on RUDDER. 
The dataset is much smaller than the others, so the retrieval performance is generally lower.
However, C2KD still improves the results across languages and metrics.
The largest improvement is on R@10 for Hindi, from 16.8 to 19.4 (+15.9\% relative).

Overall, C2KD consistently improved performance across languages and domains, with significant improvements on some languages. 
Also, our results accurately represent the performance since we ran each experiment three times and report the average.

\subsection{Qualitative Results}
Figure~\ref{fig:qual} shows qualitative retrieval results on Multi-YouCook2.
Without C2KD, the baseline method often retrieves clips that are only partially related to the query, ie., for ``add the beef and tofu to the pot,'' the clips will only show either beef or tofu.
Using C2KD, the model can handle more complex queries and retrieve clips that are relevant to all of the ingredients mentioned in the text.

\noindent \textbf{Retrieval using unseen languages.} Qualitative results show that our C2KD model can retrieve videos using text in \textit{unseen languages} (languages for which no text-video pairs were available), thanks to LaBSE's text pre-training in over 100 languages. 
For example, in our multilingual text-video retrieval demo\footnote{https://github.com/roudimit/c2kd}, our model can match videos with text in Ukrainian and Igbo, even though the model was not trained with text-video pairs in those languages (although LaBSE was pre-trained with text in those languages).
\section{Conclusion}
In this work, we introduce Cross-Lingual Cross-Modal Knowledge Distillation (C2KD) to improve multilingual text-video retrieval performance.
Motivated by the observation that English retrieval outperforms other languages, our method trains a student using input multilingual text to output similar text-video similarity scores compared with teachers using input English text.
We propose an objective based on cross entropy to distill the cross-modal knowledge from the teachers which considers the context of all of the text-video pairs in the batch.
We applied C2KD to four datasets and obtained an improvement in multilingual text-video retrieval across languages and domains.
Finally, we introduce the Multi-YouCook2 dataset with captions in 9 languages and will make the data public to spur more research in this direction.
Ideas for future work include applying multilingual text augmentation and paraphrasing strategies to generate more data.
\section*{Limitations}

In this work, we sought to improve multilingual text-video retrieval and reduce the gap with English performance.
C2KD improved multilingual text-video retrieval across datasets and languages.
On Multi-MSR-VTT, the average gap in performance between non-English languages and English was 16.5\% before applying C2KD and 14.4\% after applying C2KD.
On Multi-YouCook2, the gap was 20.8\% before and 18.4\% after.
Although our method reduced the gap, the performance for English is still higher than the other languages.
We attribute this to several factors.
First, C2KD improved the performance for English as well as for other languages, making it harder to close the gap.
Second, multilingual text translated from English often has errors.
For example, as we noted on Multi-MSRVTT, the performance for Vietnamese ($vi$) is much lower than the other languages, and we found the translations to be of poor quality.
Third, the multilingual text models such as LaBSE are pre-trained on more English data than any other language.
We expect that the gap between English performance and other languages will decrease as machine translation models and multilingual text encoders improve.
Also, our datasets have at most 9 languages, and it will take further research to develop \textit{massively} multilingual text-video retrieval.

\section*{Ethics Statement}
Text-video retrieval is an important task that can improve the experience of searching for videos on the internet.
Our approach aims to make text-video retrieval more equitable by improving the performance for more languages besides English.
We believe that our work can help both researchers and practitioners develop better multilingual text-video models.
We also collected multilingual captions for the YouCook2 dataset and plan to release them, which is permitted by YouCook2’s license (MIT license).

\section*{Acknowledgements}
This research was supported by the MIT-IBM Watson AI Lab and the MIT SuperCloud~\cite{reuther2018interactive}. We thank Alex H. Liu and Yuan Gong for helpful comments.

\bibliography{emnlp2022}

\begin{thebibliography}{43}
\expandafter\ifx\csname natexlab\endcsname\relax\def\natexlab#1{#1}\fi

\bibitem[{Akula et~al.(2021)Akula, Dabral, Jyothi, and
  Ramakrishnan}]{akula2021cross}
Jayaprakash Akula, Rishabh Dabral, Preethi Jyothi, and Ganesh Ramakrishnan.
  2021.
\newblock Cross lingual video and text retrieval: A new benchmark dataset and
  algorithm.
\newblock In \emph{Proceedings of the 2021 International Conference on
  Multimodal Interaction}, pages 595--603.

\bibitem[{Bagher~Zadeh et~al.(2020)Bagher~Zadeh, Cao, Hessner, Liang, Poria,
  and Morency}]{bagher-zadeh-etal-2020-cmu}
AmirAli Bagher~Zadeh, Yansheng Cao, Simon Hessner, Paul~Pu Liang, Soujanya
  Poria, and Louis-Philippe Morency. 2020.
\newblock \href {https://doi.org/10.18653/v1/2020.emnlp-main.141}
  {{CMU}-{MOSEAS}: A multimodal language dataset for {S}panish, {P}ortuguese,
  {G}erman and {F}rench}.
\newblock In \emph{Proceedings of the 2020 Conference on Empirical Methods in
  Natural Language Processing (EMNLP)}, pages 1801--1812, Online. Association
  for Computational Linguistics.

\bibitem[{Chen et~al.(2021)Chen, Rouditchenko, Duarte, Kuehne, Thomas, Boggust,
  Panda, Kingsbury, Feris, Harwath et~al.}]{chen2021multimodal}
Brian Chen, Andrew Rouditchenko, Kevin Duarte, Hilde Kuehne, Samuel Thomas,
  Angie Boggust, Rameswar Panda, Brian Kingsbury, Rogerio Feris, David Harwath,
  et~al. 2021.
\newblock Multimodal clustering networks for self-supervised learning from
  unlabeled videos.
\newblock In \emph{Proceedings of the IEEE/CVF International Conference on
  Computer Vision}, pages 8012--8021.

\bibitem[{Conneau et~al.(2020)Conneau, Khandelwal, Goyal, Chaudhary, Wenzek,
  Guzm{\'a}n, Grave, Ott, Zettlemoyer, and
  Stoyanov}]{conneau-etal-2020-unsupervised}
Alexis Conneau, Kartikay Khandelwal, Naman Goyal, Vishrav Chaudhary, Guillaume
  Wenzek, Francisco Guzm{\'a}n, Edouard Grave, Myle Ott, Luke Zettlemoyer, and
  Veselin Stoyanov. 2020.
\newblock \href {https://doi.org/10.18653/v1/2020.acl-main.747} {Unsupervised
  cross-lingual representation learning at scale}.
\newblock In \emph{Proceedings of the 58th Annual Meeting of the Association
  for Computational Linguistics}, pages 8440--8451, Online. Association for
  Computational Linguistics.

\bibitem[{Croitoru et~al.(2021)Croitoru, Bogolin, Leordeanu, Jin, Zisserman,
  Albanie, and Liu}]{croitoru2021teachtext}
Ioana Croitoru, Simion-Vlad Bogolin, Marius Leordeanu, Hailin Jin, Andrew
  Zisserman, Samuel Albanie, and Yang Liu. 2021.
\newblock Teachtext: Crossmodal generalized distillation for text-video
  retrieval.
\newblock In \emph{Proceedings of the IEEE/CVF International Conference on
  Computer Vision}, pages 11583--11593.

\bibitem[{Dauphin et~al.(2017)Dauphin, Fan, Auli, and
  Grangier}]{dauphin2017language}
Yann~N Dauphin, Angela Fan, Michael Auli, and David Grangier. 2017.
\newblock Language modeling with gated convolutional networks.
\newblock In \emph{International conference on machine learning}, pages
  933--941. PMLR.

\bibitem[{Devlin et~al.(2019)Devlin, Chang, Lee, and
  Toutanova}]{devlin-etal-2019-bert}
Jacob Devlin, Ming-Wei Chang, Kenton Lee, and Kristina Toutanova. 2019.
\newblock \href {https://doi.org/10.18653/v1/N19-1423} {{BERT}: Pre-training of
  deep bidirectional transformers for language understanding}.
\newblock In \emph{Proceedings of the 2019 Conference of the North {A}merican
  Chapter of the Association for Computational Linguistics: Human Language
  Technologies, Volume 1 (Long and Short Papers)}, pages 4171--4186,
  Minneapolis, Minnesota. Association for Computational Linguistics.

\bibitem[{Ephrat et~al.(2018)Ephrat, Mosseri, Lang, Dekel, Wilson, Hassidim,
  Freeman, and Rubinstein}]{10.1145/3197517.3201357}
Ariel Ephrat, Inbar Mosseri, Oran Lang, Tali Dekel, Kevin Wilson, Avinatan
  Hassidim, William~T. Freeman, and Michael Rubinstein. 2018.
\newblock \href {https://doi.org/10.1145/3197517.3201357} {Looking to listen at
  the cocktail party: A speaker-independent audio-visual model for speech
  separation}.
\newblock \emph{ACM Trans. Graph.}, 37(4).

\bibitem[{Feng et~al.(2022)Feng, Yang, Cer, Arivazhagan, and
  Wang}]{feng-etal-2022-language}
Fangxiaoyu Feng, Yinfei Yang, Daniel Cer, Naveen Arivazhagan, and Wei Wang.
  2022.
\newblock \href {https://doi.org/10.18653/v1/2022.acl-long.62}
  {Language-agnostic {BERT} sentence embedding}.
\newblock In \emph{Proceedings of the 60th Annual Meeting of the Association
  for Computational Linguistics (Volume 1: Long Papers)}, pages 878--891,
  Dublin, Ireland. Association for Computational Linguistics.

\bibitem[{Gao et~al.(2021)Gao, Yao, and Chen}]{gao-etal-2021-simcse}
Tianyu Gao, Xingcheng Yao, and Danqi Chen. 2021.
\newblock \href {https://doi.org/10.18653/v1/2021.emnlp-main.552} {{S}im{CSE}:
  Simple contrastive learning of sentence embeddings}.
\newblock In \emph{Proceedings of the 2021 Conference on Empirical Methods in
  Natural Language Processing}, pages 6894--6910, Online and Punta Cana,
  Dominican Republic. Association for Computational Linguistics.

\bibitem[{Gong et~al.(2022)Gong, Khurana, Rouditchenko, and
  Glass}]{gong2022cmkd}
Yuan Gong, Sameer Khurana, Andrew Rouditchenko, and James Glass. 2022.
\newblock Cmkd: Cnn/transformer-based cross-model knowledge distillation for
  audio classification.
\newblock \emph{arXiv preprint arXiv:2203.06760}.

\bibitem[{Gupta et~al.(2022{\natexlab{a}})Gupta, Gautam, and
  Mamidi}]{gupta2022cvil}
Kshitij Gupta, Devansh Gautam, and Radhika Mamidi. 2022{\natexlab{a}}.
\newblock cvil: Cross-lingual training of vision-language models using
  knowledge distillation.
\newblock In \emph{2022 26th International Conference on Pattern Recognition
  (ICPR)}, pages 1734--1741. IEEE.

\bibitem[{Gupta et~al.(2022{\natexlab{b}})Gupta, Mittal, Mathur, Mishra,
  Maheshwari, Bera, Mukherjee, and Manocha}]{gupta20223massiv}
Vikram Gupta, Trisha Mittal, Puneet Mathur, Vaibhav Mishra, Mayank Maheshwari,
  Aniket Bera, Debdoot Mukherjee, and Dinesh Manocha. 2022{\natexlab{b}}.
\newblock 3massiv: multilingual, multimodal and multi-aspect dataset of social
  media short videos.
\newblock In \emph{Proceedings of the IEEE/CVF Conference on Computer Vision
  and Pattern Recognition}, pages 21064--21075.

\bibitem[{Gutmann and Hyv{\"a}rinen(2010)}]{gutmann2010noise}
Michael Gutmann and Aapo Hyv{\"a}rinen. 2010.
\newblock Noise-contrastive estimation: A new estimation principle for
  unnormalized statistical models.
\newblock In \emph{AISTATS}.

\bibitem[{Hinton et~al.(2015)Hinton, Vinyals, Dean
  et~al.}]{hinton2015distilling}
Geoffrey Hinton, Oriol Vinyals, Jeff Dean, et~al. 2015.
\newblock Distilling the knowledge in a neural network.
\newblock \emph{NIPS Deep Learning Workshop}, 2(7).

\bibitem[{Huang et~al.(2021)Huang, Patrick, Hu, Neubig, Metze, and
  Hauptmann}]{huang2021multilingual}
Po-Yao Huang, Mandela Patrick, Junjie Hu, Graham Neubig, Florian Metze, and
  Alexander~G Hauptmann. 2021.
\newblock \href {https://arxiv.org/abs/2103.08849} {Multilingual multimodal
  pre-training for zero-shot cross-lingual transfer of vision-language models}.
\newblock In \emph{Proceedings of the 2021 Conference of the North American
  Chapter of the Association for Computational Linguistics: Human Language
  Technologies}, pages 2443--2459.

\bibitem[{Jozefowicz et~al.(2016)Jozefowicz, Vinyals, Schuster, Shazeer, and
  Wu}]{jozefowicz2016exploring}
Rafal Jozefowicz, Oriol Vinyals, Mike Schuster, Noam Shazeer, and Yonghui Wu.
  2016.
\newblock Exploring the limits of language modeling.
\newblock \emph{arXiv preprint arXiv:1602.02410}.

\bibitem[{Kay et~al.(2017)Kay, Carreira, Simonyan, Zhang, Hillier,
  Vijayanarasimhan, Viola, Green, Back, Natsev et~al.}]{kay2017kinetics}
Will Kay, Joao Carreira, Karen Simonyan, Brian Zhang, Chloe Hillier, Sudheendra
  Vijayanarasimhan, Fabio Viola, Tim Green, Trevor Back, Paul Natsev, et~al.
  2017.
\newblock The kinetics human action video dataset.
\newblock \emph{arXiv preprint arXiv:1705.06950}.

\bibitem[{Kingma and Ba(2015)}]{kingma2015adam}
Diederik~P Kingma and Jimmy Ba. 2015.
\newblock Adam: A method for stochastic optimization.
\newblock In \emph{ICLR}.

\bibitem[{Lei et~al.(2021)Lei, Berg, and Bansal}]{lei-etal-2021-mtvr}
Jie Lei, Tamara Berg, and Mohit Bansal. 2021.
\newblock \href {https://doi.org/10.18653/v1/2021.acl-short.92} {m{TVR}:
  Multilingual moment retrieval in videos}.
\newblock In \emph{Proceedings of the 59th Annual Meeting of the Association
  for Computational Linguistics and the 11th International Joint Conference on
  Natural Language Processing (Volume 2: Short Papers)}, pages 726--734,
  Online. Association for Computational Linguistics.

\bibitem[{Liu et~al.(2022)Liu, Jin, Lai, Rouditchenko, Oliva, and
  Glass}]{liu-etal-2022-cross}
Alexander Liu, SouYoung Jin, Cheng-I Lai, Andrew Rouditchenko, Aude Oliva, and
  James Glass. 2022.
\newblock \href {https://doi.org/10.18653/v1/2022.acl-long.215} {Cross-modal
  discrete representation learning}.
\newblock In \emph{Proceedings of the 60th Annual Meeting of the Association
  for Computational Linguistics (Volume 1: Long Papers)}, pages 3013--3035,
  Dublin, Ireland. Association for Computational Linguistics.

\bibitem[{Madasu et~al.(2023)Madasu, Aflalo, Ben Melech~Stan, Tseng, Bertasius,
  and Lal}]{madasu2023improving}
Avinash Madasu, Estelle Aflalo, Gabriela Ben Melech~Stan, Shao-Yen Tseng, Gedas
  Bertasius, and Vasudev Lal. 2023.
\newblock Improving video retrieval using multilingual knowledge transfer.
\newblock In \emph{Advances in Information Retrieval: 45th European Conference
  on Information Retrieval, ECIR 2023, Dublin, Ireland, April 2--6, 2023,
  Proceedings, Part I}, pages 669--684. Springer.

\bibitem[{Miech et~al.(2017)Miech, Laptev, and Sivic}]{miech2017learnable}
Antoine Miech, Ivan Laptev, and Josef Sivic. 2017.
\newblock Learnable pooling with context gating for video classification.
\newblock \emph{arXiv preprint arXiv:1706.06905}.

\bibitem[{Miech et~al.(2019)Miech, Zhukov, Alayrac, Tapaswi, Laptev, and
  Sivic}]{miech2019howto100m}
Antoine Miech, Dimitri Zhukov, Jean-Baptiste Alayrac, Makarand Tapaswi, Ivan
  Laptev, and Josef Sivic. 2019.
\newblock Howto100m: Learning a text-video embedding by watching hundred
  million narrated video clips.
\newblock In \emph{Proceedings of the IEEE/CVF International Conference on
  Computer Vision}, pages 2630--2640.

\bibitem[{Oord et~al.(2018)Oord, Li, and Vinyals}]{oord2018representation}
Aaron van~den Oord, Yazhe Li, and Oriol Vinyals. 2018.
\newblock Representation learning with contrastive predictive coding.
\newblock \emph{arXiv preprint arXiv:1807.03748}.

\bibitem[{Paszke et~al.(2019)Paszke, Gross, Massa, Lerer, Bradbury, Chanan,
  Killeen, Lin, Gimelshein, Antiga, Desmaison, Kopf, Yang, DeVito, Raison,
  Tejani, Chilamkurthy, Steiner, Fang, Bai, and Chintala}]{NEURIPS2019_9015}
Adam Paszke, Sam Gross, Francisco Massa, Adam Lerer, James Bradbury, Gregory
  Chanan, Trevor Killeen, Zeming Lin, Natalia Gimelshein, Luca Antiga, Alban
  Desmaison, Andreas Kopf, Edward Yang, Zachary DeVito, Martin Raison, Alykhan
  Tejani, Sasank Chilamkurthy, Benoit Steiner, Lu~Fang, Junjie Bai, and Soumith
  Chintala. 2019.
\newblock \href
  {http://papers.neurips.cc/paper/9015-pytorch-an-imperative-style-high-performance-deep-learning-library.pdf}
  {Pytorch: An imperative style, high-performance deep learning library}.
\newblock In H.~Wallach, H.~Larochelle, A.~Beygelzimer, F.~d\textquotesingle
  Alch\'{e}-Buc, E.~Fox, and R.~Garnett, editors, \emph{Advances in Neural
  Information Processing Systems 32}, pages 8024--8035. Curran Associates, Inc.

\bibitem[{Patrick et~al.(2021)Patrick, Huang, Asano, Metze, Hauptmann,
  Henriques, and Vedaldi}]{patrick2020support}
Mandela Patrick, Po-Yao Huang, Yuki Asano, Florian Metze, Alexander Hauptmann,
  Joao Henriques, and Andrea Vedaldi. 2021.
\newblock Support-set bottlenecks for video-text representation learning.
\newblock \emph{ICLR}.

\bibitem[{Radford et~al.(2021)Radford, Kim, Hallacy, Ramesh, Goh, Agarwal,
  Sastry, Askell, Mishkin, Clark et~al.}]{radford2021learning}
Alec Radford, Jong~Wook Kim, Chris Hallacy, Aditya Ramesh, Gabriel Goh,
  Sandhini Agarwal, Girish Sastry, Amanda Askell, Pamela Mishkin, Jack Clark,
  et~al. 2021.
\newblock Learning transferable visual models from natural language
  supervision.
\newblock In \emph{International Conference on Machine Learning}, pages
  8748--8763. PMLR.

\bibitem[{Raj~Khan et~al.(2021)Raj~Khan, Gupta, and
  Ekbal}]{raj-khan-etal-2021-towards-developing}
Humair Raj~Khan, Deepak Gupta, and Asif Ekbal. 2021.
\newblock \href {https://doi.org/10.18653/v1/2021.findings-emnlp.151} {Towards
  developing a multilingual and code-mixed visual question answering system by
  knowledge distillation}.
\newblock In \emph{Findings of the Association for Computational Linguistics:
  EMNLP 2021}, pages 1753--1767, Punta Cana, Dominican Republic. Association
  for Computational Linguistics.

\bibitem[{Reimers and Gurevych(2019)}]{reimers-gurevych-2019-sentence}
Nils Reimers and Iryna Gurevych. 2019.
\newblock \href {https://doi.org/10.18653/v1/D19-1410} {Sentence-{BERT}:
  Sentence embeddings using {S}iamese {BERT}-networks}.
\newblock In \emph{Proceedings of the 2019 Conference on Empirical Methods in
  Natural Language Processing and the 9th International Joint Conference on
  Natural Language Processing (EMNLP-IJCNLP)}, pages 3982--3992, Hong Kong,
  China. Association for Computational Linguistics.

\bibitem[{Reimers and Gurevych(2020)}]{reimers-gurevych-2020-making}
Nils Reimers and Iryna Gurevych. 2020.
\newblock \href {https://doi.org/10.18653/v1/2020.emnlp-main.365} {Making
  monolingual sentence embeddings multilingual using knowledge distillation}.
\newblock In \emph{Proceedings of the 2020 Conference on Empirical Methods in
  Natural Language Processing (EMNLP)}, pages 4512--4525, Online. Association
  for Computational Linguistics.

\bibitem[{Reuther et~al.(2018)Reuther, Kepner, Byun, Samsi, Arcand, Bestor,
  Bergeron, Gadepally, Houle, Hubbell, Jones, Klein, Milechin, Mullen, Prout,
  Rosa, Yee, and Michaleas}]{reuther2018interactive}
Albert Reuther, Jeremy Kepner, Chansup Byun, Siddharth Samsi, William Arcand,
  David Bestor, Bill Bergeron, Vijay Gadepally, Michael Houle, Matthew Hubbell,
  Michael Jones, Anna Klein, Lauren Milechin, Julia Mullen, Andrew Prout,
  Antonio Rosa, Charles Yee, and Peter Michaleas. 2018.
\newblock Interactive supercomputing on 40,000 cores for machine learning and
  data analysis.
\newblock In \emph{2018 IEEE High Performance extreme Computing Conference
  (HPEC)}, pages 1--6. IEEE.

\bibitem[{Rouditchenko et~al.(2021{\natexlab{a}})Rouditchenko, Boggust,
  Harwath, Chen, Joshi, Thomas, Audhkhasi, Kuehne, Panda, Feris, Kingsbury,
  Picheny, Torralba, and Glass}]{rouditchenko21_interspeech}
Andrew Rouditchenko, Angie Boggust, David Harwath, Brian Chen, Dhiraj Joshi,
  Samuel Thomas, Kartik Audhkhasi, Hilde Kuehne, Rameswar Panda, Rogerio Feris,
  Brian Kingsbury, Michael Picheny, Antonio Torralba, and James Glass.
  2021{\natexlab{a}}.
\newblock \href {https://doi.org/10.21437/Interspeech.2021-1312} {{AVLnet:
  Learning Audio-Visual Language Representations from Instructional Videos}}.
\newblock In \emph{Interspeech}, pages 1584--1588.

\bibitem[{Rouditchenko et~al.(2021{\natexlab{b}})Rouditchenko, Boggust,
  Harwath, Thomas, Kuehne, Chen, Panda, Feris, Kingsbury, Picheny, and
  Glass}]{rouditchenko21b_interspeech}
Andrew Rouditchenko, Angie Boggust, David Harwath, Samuel Thomas, Hilde Kuehne,
  Brian Chen, Rameswar Panda, Rogerio Feris, Brian Kingsbury, Michael Picheny,
  and James Glass. 2021{\natexlab{b}}.
\newblock \href {https://doi.org/10.21437/Interspeech.2021-1352} {{Cascaded
  Multilingual Audio-Visual Learning from Videos}}.
\newblock In \emph{Proc. Interspeech 2021}, pages 3006--3010.

\bibitem[{Sanabria et~al.(2018)Sanabria, Caglayan, Palaskar, Elliott, Barrault,
  Specia, and Metze}]{sanabria2018how2}
Ramon Sanabria, Ozan Caglayan, Shruti Palaskar, Desmond Elliott, Lo\"ic
  Barrault, Lucia Specia, and Florian Metze. 2018.
\newblock {How2:} a large-scale dataset for multimodal language understanding.
\newblock In \emph{Workshop on Visually Grounded Interaction and Language
  (ViGIL)}. NeurIPS.

\bibitem[{Shvetsova et~al.(2022)Shvetsova, Chen, Rouditchenko, Thomas,
  Kingsbury, Feris, Harwath, Glass, and Kuehne}]{shvetsova2022everything}
Nina Shvetsova, Brian Chen, Andrew Rouditchenko, Samuel Thomas, Brian
  Kingsbury, Rogerio~S Feris, David Harwath, James Glass, and Hilde Kuehne.
  2022.
\newblock Everything at once-multi-modal fusion transformer for video
  retrieval.
\newblock In \emph{Proceedings of the IEEE/CVF Conference on Computer Vision
  and Pattern Recognition}, pages 20020--20029.

\bibitem[{Sigurdsson et~al.(2020)Sigurdsson, Alayrac, Nematzadeh, Smaira,
  Malinowski, Carreira, Blunsom, and Zisserman}]{sigurdsson2020visual}
Gunnar~A Sigurdsson, Jean-Baptiste Alayrac, Aida Nematzadeh, Lucas Smaira,
  Mateusz Malinowski, Joao Carreira, Phil Blunsom, and Andrew Zisserman. 2020.
\newblock Visual grounding in video for unsupervised word translation.
\newblock In \emph{Proceedings of the IEEE/CVF Conference on Computer Vision
  and Pattern Recognition}, pages 10850--10859.

\bibitem[{Su et~al.(2021)Su, Duan, Cui, Ji, Wu, Luo, Liu, Zhong, Bharti, and
  Sacheti}]{su-etal-2021-gem}
Lin Su, Nan Duan, Edward Cui, Lei Ji, Chenfei Wu, Huaishao Luo, Yongfei Liu,
  Ming Zhong, Taroon Bharti, and Arun Sacheti. 2021.
\newblock \href {https://doi.org/10.18653/v1/2021.findings-acl.229} {{GEM}: A
  general evaluation benchmark for multimodal tasks}.
\newblock In \emph{Findings of the Association for Computational Linguistics:
  ACL-IJCNLP 2021}, pages 2594--2603, Online. Association for Computational
  Linguistics.

\bibitem[{Sun et~al.(2019)Sun, Baradel, Murphy, and Schmid}]{sun2019learning}
Chen Sun, Fabien Baradel, Kevin Murphy, and Cordelia Schmid. 2019.
\newblock Learning video representations using contrastive bidirectional
  transformer.
\newblock \emph{arXiv preprint arXiv:1906.05743}.

\bibitem[{Vaswani et~al.(2017)Vaswani, Shazeer, Parmar, Uszkoreit, Jones,
  Gomez, Kaiser, and Polosukhin}]{vaswani2017attention}
Ashish Vaswani, Noam Shazeer, Niki Parmar, Jakob Uszkoreit, Llion Jones,
  Aidan~N Gomez, {\L}ukasz Kaiser, and Illia Polosukhin. 2017.
\newblock Attention is all you need.
\newblock \emph{Advances in neural information processing systems}, 30.

\bibitem[{Wang et~al.(2019)Wang, Wu, Chen, Li, Wang, and Wang}]{wang2019vatex}
Xin Wang, Jiawei Wu, Junkun Chen, Lei Li, Yuan-Fang Wang, and William~Yang
  Wang. 2019.
\newblock Vatex: A large-scale, high-quality multilingual dataset for
  video-and-language research.
\newblock In \emph{Proceedings of the IEEE/CVF International Conference on
  Computer Vision}, pages 4581--4591.

\bibitem[{Xu et~al.(2016)Xu, Mei, Yao, and Rui}]{xu2016msr}
Jun Xu, Tao Mei, Ting Yao, and Yong Rui. 2016.
\newblock Msr-vtt: A large video description dataset for bridging video and
  language.
\newblock In \emph{Proceedings of the IEEE conference on computer vision and
  pattern recognition}, pages 5288--5296.

\bibitem[{Zhou et~al.(2018)Zhou, Xu, and Corso}]{zhou2018towards}
Luowei Zhou, Chenliang Xu, and Jason~J Corso. 2018.
\newblock Towards automatic learning of procedures from web instructional
  videos.
\newblock In \emph{Thirty-Second AAAI Conference on Artificial Intelligence}.

\end{thebibliography}
\bibliographystyle{acl_natbib}

\appendix
\section{Appendix}
\label{sec:appendix}

Table~\ref{tab:params} shows the final values of the $\alpha$ (balance) and $\Psi$ (pooler) hyperparameters determined via grid-search.
As shown by the table, a larger weight on $\mathcal{L}_{C2KD}$ is optimal for most of the datasets.
We have also tried a multitask setup where no pooler is used and $\mathcal{L}_{C2KD}$ is applied to each teacher text-video similarity matrix, but this approach did not work as well.

\begin{table}[h]
\small
\centering
\resizebox{0.48\textwidth}{!}{%
\begin{tabular}{lcccc}\hline
    Parameter & MSRVTT & YouCook2 & VATEX & RUDDER  \\
    \hline\hline
    $\alpha$ (Balance) & 0.5 & 0.1 & 0.1 & 0.1 \\
    $\Psi$ (Pooler) & Min & Min & Max & Mean \\
    \hline
\end{tabular}}
\caption{Final values of the balance and pooler hyperparameters determined via grid-search.}
\label{tab:params}
\end{table}

For the video model, we use a maximum video length of 30s.
Following the 2-layer, 4-head transformer, we mean-pool the outputs and apply a projection into the shared embedding space with a dimension of 512.
Following prior work in text-video learning, we use non-linear feature gating in the projection layer~\cite{dauphin2017language,miech2017learnable} and we do not use positional embeddings in the video transformer~\cite{shvetsova2022everything}.

We used the HuggingFace models and tokenizers for LaBSE\footnote{https://huggingface.co/sentence-transformers/LaBSE}, XLM-R\footnote{https://huggingface.co/xlm-roberta-base}, mBERT\footnote{https://huggingface.co/bert-base-multilingual-uncased}, Distill SBERT\footnote{https://huggingface.co/sentence-transformers/distiluse-base-multilingual-cased-v2}, and SimCSE\footnote{https://huggingface.co/princeton-nlp/sup-simcse-roberta-base}.
We use a maximum of 40 tokens.
Following the text encoders, we mean-pool the outputs and apply a projection into the shared embedding space.

We used a single V100 GPU with 32 GB memory for all of our experiments, and each experiment took one hour on average.
We trained the models for 20 epochs for MSR-VTT, 10 epochs for Multi-YouCook2, 30 epochs for VATEX, and 20 epochs for RUDDER. 
The batch size was 64 videos.
The initial learning rate was 1e-4 with an exponential decay of 0.9 except on RUDDER where we reduced it to 5e-5 due to the smaller dataset size.
Models were trained with the Adam optimizer~\cite{kingma2015adam}.
We implemented the models in PyTorch~\cite{NEURIPS2019_9015} and used mixed-precision.

\end{document}